\documentclass{article}
\usepackage{spconf,graphicx}
\usepackage{amsmath}
\usepackage{amssymb}
\usepackage{mathtools}
\usepackage[euler]{textgreek}
\usepackage{color}
\usepackage{soul}

\usepackage{makeidx}
\usepackage[scriptsize,center,tight]{subfigure}
\usepackage{setspace}
\usepackage{xspace}
\usepackage{array}

\usepackage[percent]{overpic}

\usepackage{siunitx}
\usepackage{xcolor}

\usepackage{enumitem}
\setlist{nosep, leftmargin=14pt}

\usepackage{mwe} 

\usepackage{eso-pic}

\title{Deterministic Temporal Embedding for Calcium Imaging Trace Extraction}
\name{
\fontsize{10.5}{12.5}\selectfont
Qi Gao$^{1,2}$, Julius Beichert$^{3,4}$, Miriam Ratliff$\,^{3,4}$, and Karl Rohr$^2$
}

\address{
\begin{minipage}{0.98\textwidth}
\centering
\fontsize{9}{10.5}\selectfont
$^{1}$Core Facility Platform Mannheim,
CF LIMa, Medical Faculty Mannheim,
Heidelberg University, Germany\\
$^{2}$Biomedical Computer Vision Group,
BioQuant, IPMB, Heidelberg University, Germany\\
$^{3}$Department of Neurosurgery,
University Hospital Mannheim,
Heidelberg University, Germany\\
$^{4}$Clinical Cooperation Unit Neurooncology,
German Cancer Consortium (DKTK),
German Cancer Research Center (DKFZ), Germany
\end{minipage}
}

\newcommand{\myparagraph}[1]{\vspace{0.3\baselineskip}\noindent\textbf{#1}}

\newcommand{\Eq}{Eq.\@\xspace}

\newcommand{\Fig}{Fig.\@\xspace}

\newcommand{\Tab}{Tab.\@\xspace}

\def\onedot{\ifx\@let@token.\else.\null\fi\xspace}
\def\eg{\emph{e.g}\onedot}

\def\cf{\emph{cf}\onedot}

\def\wrt{w.r.t\onedot}

\def\d{\ensuremath{d}}

\newcommand{\R}{\ensuremath{\mathbb{R}}}

\newcommand{\invisible}[1]{}

\newcommand\y{\mathbf{y}}

\renewcommand\d{\mathbf{d}}
\newcommand\f{\mathbf{f}}

\renewcommand\k{\mathbf{k}}

\newcommand\m{\mathbf{m}}

\newcolumntype{P}[1]{>{\centering\arraybackslash}p{#1}}

\begin{document}

\AddToShipoutPictureFG*{
\AtPageLowerLeft{
\raisebox{5mm}{
\makebox[\paperwidth][c]{
\parbox{0.9\paperwidth}{
\centering\scriptsize
This work has been submitted to the IEEE for possible publication.
Copyright may be transferred without notice, after which this version
may no longer be accessible.}}}}}

\pagestyle{plain}

%
\maketitle
\begin{abstract}
Reliable extraction of calcium-dependent fluorescence traces in calcium imaging 
sequences requires identifying pixels whose fluorescence intensity reflects the same 
underlying cellular activity. 
We present a deterministic temporal embedding framework that maps pixel-wise 
fluorescence traces to a low-dimensional representation in which temporally 
coherent sources form spatially coherent regions. 
Starting from a fuzzy-neighborhood graph and spectral initialization, we optimize a 
class-balanced cross-entropy objective using exact sparse attraction and a deterministic 
particle-mesh decomposition of the dense repulsive interaction with sparse edge 
compensation, avoiding stochastic negative sampling. 
The resulting latent representation is spatially decoded into reliable source supports 
for trace extraction without imposing an explicit cellular shape prior. 
Experiments on synthetic and in vivo two-photon calcium imaging data demonstrate 
robust recovery of irregular and spatially touching sources together with efficient 
deterministic embedding optimization.
\end{abstract}
\begin{keywords}
Calcium Imaging, Trace Extraction, Embedding
\end{keywords}

\section{Introduction}
\label{sec:intro}

Calcium imaging enables the observation of intracellular calcium dynamics across 
large populations of cells and provides a widely used readout of cellular activity. 
Downstream biological analysis requires determining a reliable fluorescence or 
calcium activity trace associated with each underlying cellular source. 
Accurate and reproducible source support estimation for trace extraction is therefore 
a central step from raw calcium imaging data to quantitative analysis of cellular 
dynamics.

Existing calcium imaging trace extraction methods can generally be classified into 
three categories. 
First, widely used ROI- and clustering-based methods identify active 
cellular regions and subsequently extract their temporal signals (\eg, 
PCA/ICA-based cell sorting~\cite{mukamel2009automated}, 
Suite2p~\cite{stringer2026suite2p}, and MIN1PIPE~\cite{lu2018min1pipe}). 
A second category of methods formulates source extraction as a spatio-temporal 
decomposition problem. 
Constrained nonnegative matrix factorization (CNMF)~\cite{pnevmatikakis2016CNMF}, 
CNMF-E~\cite{zhou2018CNMFE}, and CaImAn \cite{giovannucci2019caiman} jointly 
estimate spatial footprints, temporal components, and background activity, and are 
particularly effective when their source and background models match the imaging 
modality. 
Third, more recently, deep learning methods have been used to learn spatio-temporal 
features for neuron detection and segmentation directly from calcium imaging 
sequences (\eg, STNeuroNet~\cite{soltanian2019fast}, SUNS~\cite{bao2023SUNS}, 
and DeepWonder~\cite{zhang2023deepwonder}). 
These methods have improved automation, scalability, and segmentation accuracy, 
but they generally rely on spatial source models, on morphology learned from 
annotated or simulated neurons, or on assumptions tailored to commonly studied 
neuronal imaging settings.

\begin{figure}[t]
\centering \hfill
\subfigcapskip=-2pt
\subfigure[Frame 20]{
\includegraphics[width=0.38\columnwidth]{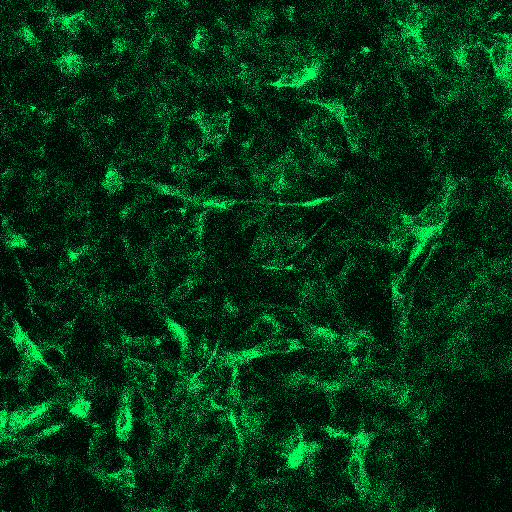}} \hfill
\subfigure[Frame 110]{
\includegraphics[width=0.38\columnwidth]{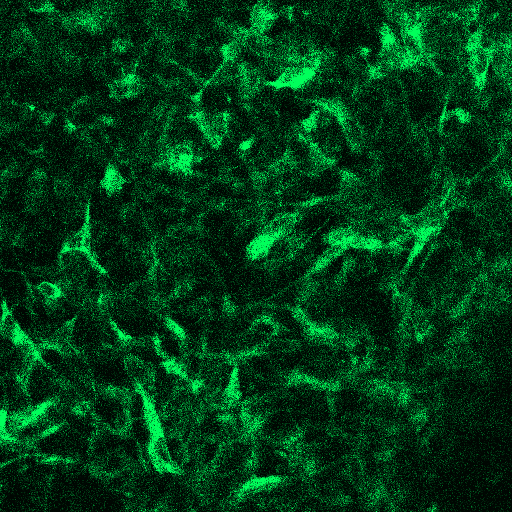}} \hspace*{\fill} \\
\vspace{-6pt}
\hfill
\subfigure[UMAP]{
\label{fig:umap}
\includegraphics[width=0.38\columnwidth]{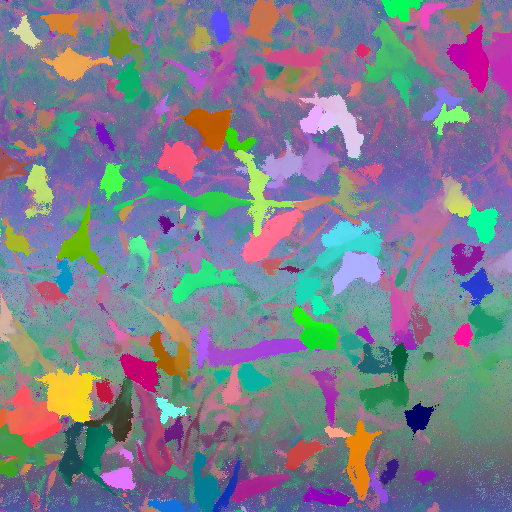}} \hfill
\subfigure[Deterministic PM]{
\label{fig:pm}
\includegraphics[width=0.38\columnwidth]{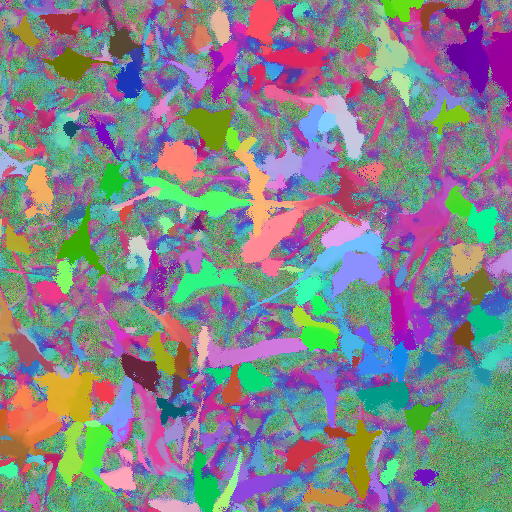}} 
\vspace{-6pt}
\hspace*{\fill} 
\caption{\textit{(Top)} Example frames from an in vivo two-photon calcium imaging 
sequence of glioblastoma cells; 
\textit{(Bottom)} 3D temporal embeddings of the entire sequence generated 
using UMAP and our deterministic particle-mesh (PM) method, 
with latent coordinates mapped to RGB for visualization.}
\label{fig:img_umap_pm}
\vspace{-3pt}
\end{figure}

Here, we consider two-photon calcium imaging data of glioblastoma cells, in which 
active cellular sources may be irregular, elongated, branching, curved, heterogeneous 
in brightness, and frequently touching or spatially connected 
(\cf \Fig~\ref{fig:img_umap_pm}(a,b)). 
In such data, a static or maximum-intensity projection may merge distinct cells, 
and morphology alone is generally insufficient to define source identity.
In contrast, pixels belonging to the same active cell often exhibit a common blinking 
pattern over time, whereas neighboring cells can display distinct and asynchronous 
temporal dynamics. 
This motivates treating source identification as a high-dimensional temporal 
representation problem rather than as a purely spatial image segmentation problem. 
In our preliminary work, we noticed that UMAP~\cite{mcinnes2018umap} applied 
directly to raw pixel-wise fluorescence traces produced remarkably compact latent 
regions corresponding to blinking cells, with neighboring cells often mapped to distinct 
latent values even when being in direct spatial contact (\cf \Fig~\ref{fig:umap}). 
However, standard UMAP uses stochastic low-dimensional optimization, which is 
undesirable when the embedding itself serves as the basis for reproducible downstream 
source extraction.

In this work, we introduce a deterministic temporal embedding framework inspired by 
UMAP for calcium imaging trace extraction. 
Pixel-wise raw fluorescence traces~are represented by complementary Euclidean and 
cosine fuzzy-neighborhood graphs, which are adaptively fused to retain both waveform 
and amplitude information. 
Starting from a spectral initialization used in UMAP, we optimize a class-balanced 
cross-entropy objective in which sparse graph attraction is computed exactly, while 
the dense repulsive interaction is decomposed into a global particle-mesh (PM) field 
and an exact sparse edge compensation term. 
This formulation avoids stochastic negative sampling and enables deterministic 
low-dimensional optimization. 
The resulting latent representation is decoded into spatially coherent candidate source 
supports, which are then selected based on temporal activity, without imposing an 
explicit cellular shape prior. 
Experiments on synthetic data with ground truth and real in vivo calcium imaging 
data demonstrate robust recovery of irregular and spatially touching sources 
and reliable extraction of their fluorescence traces.

\section{Methods}
\label{sec:methods}

\subsection{Temporal fuzzy graph and spectral initialization}
\vspace{-3pt}

We start from the raw fluorescence trace of each spatial pixel in an image 
sequence of $T$ frames.
No temporal centering or spatial smoothing is applied, as both operations can 
suppress or mix source-specific temporal structure.
We construct fuzzy neighborhood graphs from the raw traces using the 
standard UMAP neighborhood construction~\cite{mcinnes2018umap} 
with 15 nearest neighbors.
Two graphs are computed independently, using Euclidean and cosine 
distance.
The Euclidean graph retains amplitude-dependent information, whereas 
the cosine graph emphasizes waveform similarity and is less sensitive to 
multiplicative intensity variation within a source.
The proposed representation combines these complementary relationships.
Let $w_{ij}^{E}$ and $w_{ij}^{C}$ denote the Euclidean and cosine fuzzy
membership strengths between pixels $i$ and $j$, respectively.
We first compute
\vspace{-1pt}
\begin{equation}
\tilde{w}_{ij} = 
\alpha\max\!\left(w_{ij}^E,w_{ij}^C\right) 
+ (1\!-\!\alpha)\min\!\left(w_{ij}^{E},w_{ij}^{C}\right),
\label{eq:fusion}
\vspace{-1pt}
\end{equation}
with $\alpha=0.75$.
Thus, a strong relationship in either metric is retained, while agreement 
between the two metrics provides additional support.
The fused graph is subsequently sparsified to the 15 strongest memberships 
per node and then symmetrized with the fuzzy union
$w_{ij}=\tilde{w}_{ij}+\tilde{w}_{ji}-\tilde{w}_{ij}\tilde{w}_{ji}$.

Following the spectral initialization used in UMAP, we use a $D$-dimensional 
spectral embedding of the fused fuzzy graph to initialize the latent coordinates. 
This provides a reproducible coarse graph organization before the subsequent 
low-dimensional optimization and avoids introducing random large-scale 
structure at initialization.

\vspace{-3pt}
\subsection{Deterministic particle-mesh temporal embedding}
\vspace{-3pt}

While the theoretical UMAP objective compares high- and low-dimensional 
fuzzy memberships through a binary cross-entropy, a sampling-based 
objective is optimized in practice.
As shown in \cite{damrich2021umap}, this sampling procedure substantially 
changes the effective loss and, importantly, implicitly balances the sparse 
attractive interactions against the much larger population of repulsive pairs. 

We use $\y_i\!\in\!\R^D$ to denote the latent coordinate of pixel~$i$, and 
define the displacement $\d_{ij}\!=\!\y_i\!-\!\y_j$.
Using the low-dimensional fuzzy membership 
$q_{ij} \!=\! 1/(1+a\|\d_{ij}\|^{2b})$ and the curve parameters $a$ and $b$ 
as in UMAP, we define the class-balanced cross-entropy objective
\vspace{-1pt}
\begin{equation}
\vspace{-6pt}
\mathcal{L} = - \frac{1}{2} \sum_i \sum_{j\ne i} \Big ( w_{ij} \!\log q_{ij} 
+ \beta \lambda (1\!-w_{ij}\!) \log(1\!-q_{ij}\!) \Big ),
\label{eq:loss}
\end{equation}
where 
\vspace{-3pt}
\begin{equation}
\vspace{-1pt}
\lambda = \tfrac{\sum_i\sum_{j\ne i}w_{ij}}{\sum_i\sum_{j\ne i}(1-w_{ij})}, 
\end{equation}
and $\beta\!>\!0$ is a global weighting parameter.
The two terms in the negative gradient of the loss \wrt $\y_i$ are named 
attractive and repulsive forces
\vspace{-3pt}
\begin{equation}
\vspace{-3pt}
- \nabla_{\y_i}\mathcal{L} = \f_i^{\mathrm{att}} + \beta\f_i^{\mathrm{rep}}.
\end{equation}
The attractive force is computed over the neighborhood $\mathcal{N}_i$ of 
pixel $i$ in the sparse fused fuzzy graph
\vspace{-3pt}
\begin{equation}
\vspace{-3pt}
\f_i^{\mathrm{att}} = - \sum\nolimits_{j\in\mathcal{N}_i} w_{ij}
\frac{2ab\|\d_{ij}\|^{2(b-1)}}{1+a\|\d_{ij}\|^{2b}} \d_{ij}.
\end{equation}
By defining a repulsive kernel for a displacement $\mathbf{d}$ as
\vspace{-3pt}
\begin{equation}
\vspace{-3pt}
\mathbf{k}_{\mathrm{r}}(\d) =
\frac{2b}{\left(\epsilon+\|\d\|^2\right)  \left(1+a\|\d\|^{2b}\right)} \d,
\end{equation}
with $\epsilon\!=\!10^{-3}$ for numerical stability, the repulsive force is 
\vspace{-3pt}
\begin{equation}
\vspace{-3pt}
\begin{aligned}[b]
\f_i^{\mathrm{rep}} 
& = \lambda \sum\nolimits_{j\ne i}(1-w_{ij})\k_{\mathrm{r}}(\d_{ij}) \\
&  = \lambda \left( \sum\nolimits_{j\ne i} \k_{\mathrm{r}}(\d_{ij}) 
- \sum\nolimits_{j\in\mathcal{N}_i} w_{ij} \k_{\mathrm{r}}(\d_{ij}) \right).
\end{aligned}
\end{equation}
While the second term (edge compensation) is computed exactly on the sparse 
fuzzy graph, the first term (global repulsion) requires all pairwise interactions and 
must be carefully treated.

To efficiently compute the global repulsion, the latent space is represented 
on a $D$-dimensional Cartesian mesh.
Each latent point is treated as a unit-mass particle and deposited onto its 
$2^D$ neighboring mesh vertices using cloud-in-cell interpolation.
If $\omega_{ig}$ denotes the interpolation weight of particle $i$ at mesh 
vertex $g\!\in\!\mathcal{G}$, with $\mathcal{G}$ denoting the mesh vertex set, 
the resulting scalar particle-density field is
\vspace{-3pt}
\begin{equation}
\vspace{-3pt}
\rho_g = \sum\nolimits_i \omega_{ig}.
\end{equation}
The repulsive vector field on the mesh is then
\vspace{-3pt}
\begin{equation}
\vspace{-3pt}
\mathbf{r}_g = \sum\nolimits_{g^\prime\in\mathcal{G}} 
\rho_{g^\prime} \mathbf{k}_{\mathrm{r}}(\m_g-\m_{g^\prime})
= (\rho * \k_{\mathrm{r}})_g,
\end{equation}
where $\m_g\!\in\!\R^D$ denotes the coordinate of mesh vertex $g$.
This convolution can be computed efficiently using the FFT-based particle-mesh 
scheme described in~\cite{linderman2019fitsne}.
Since a latent particle generally does not coincide with a mesh vertex, the global 
repulsion term is approximated by interpolating the mesh field back to the particle
\vspace{-3pt}
\begin{equation}
\vspace{-3pt}
\sum\nolimits_{j\ne i} \k_{\mathrm{r}}(\d_{ij}) 
\approx \sum\nolimits_g \omega_{ig} \mathbf{r}_g.
\end{equation}

The synchronous embedding update is
\vspace{-3pt}
\begin{equation}
\vspace{-3pt}
\y_{i}^{(t+1)} = \y_{i}^{(t)} + \eta_t 
\left( \f_i^{\mathrm{att}} + \beta \f_i^{\mathrm{rep}} \right),
\end{equation}
where $\eta_t$ is the step size.
The embedding is recentered after each update.
Given a fixed fuzzy graph and spectral initialization, all particle-mesh 
optimization steps are deterministic.

In contrast to~\cite{chen2026ibumap}, in which a repulsive field is constructed 
to mimic the aggregate behavior of UMAP's stochastic optimizer, 
our PM method is based on the explicit objective in \Eq~\eqref{eq:loss}, with 
the attraction--repulsion structure and PM--sparse decomposition analytically 
derived from it.

\vspace{-3pt}
\subsection{Spatial support and trace extraction}
\vspace{-3pt}

The $D$-dimensional latent coordinates are mapped back to their image 
locations to form a spatial latent field.
Since the resulting field is typically smooth within sources but changes 
sharply across source boundaries (\cf~\Fig~\ref{fig:pm}),
useful supports can be recovered with a relatively simple spatial decoder.
In this work, we use mean-shift segmentation~\cite{comaniciu2002meanshift} 
followed by merging of adjacent regions with similar latent representations, 
as one practical implementation rather than a core component of the 
embedding method.
The specific decoder can be replaced by other segmentation methods.

Candidate supports are subsequently selected and ranked using a short-lag
positive autocorrelation (ACF) activity feature, motivated by previous use 
in calcium imaging to distinguish signal-bearing
components from noise~\cite{weiser2023segmentation}.
The ACF is used only as a post-embedding activity measure and is not part 
of the embedding optimization.
Finally, the fluorescence trace of each accepted source is obtained as the
frame-wise mean intensity over its selected support.

\section{Experiments}
\label{sec:exp}
\vspace{-3pt}

\begin{figure*}[htbp]
\centering \hfill
\subfigure{ \label{fig:syn_results_a}
\begin{overpic}[width=0.3\columnwidth]{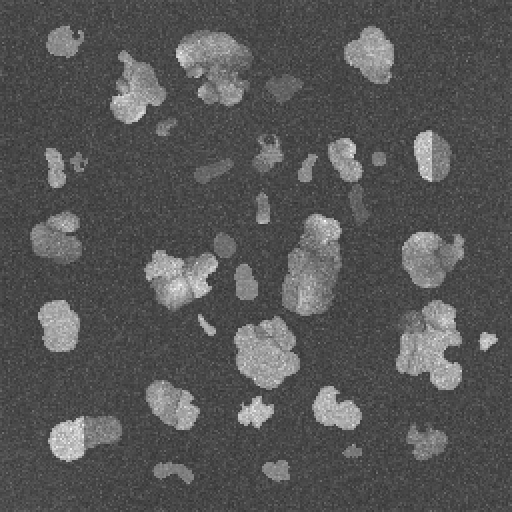} 
\put(2,2){\textcolor{white}{\footnotesize \bfseries (a)}} 
\end{overpic}
} \hfill
\subfigure{
\begin{overpic}[width=0.3\columnwidth]{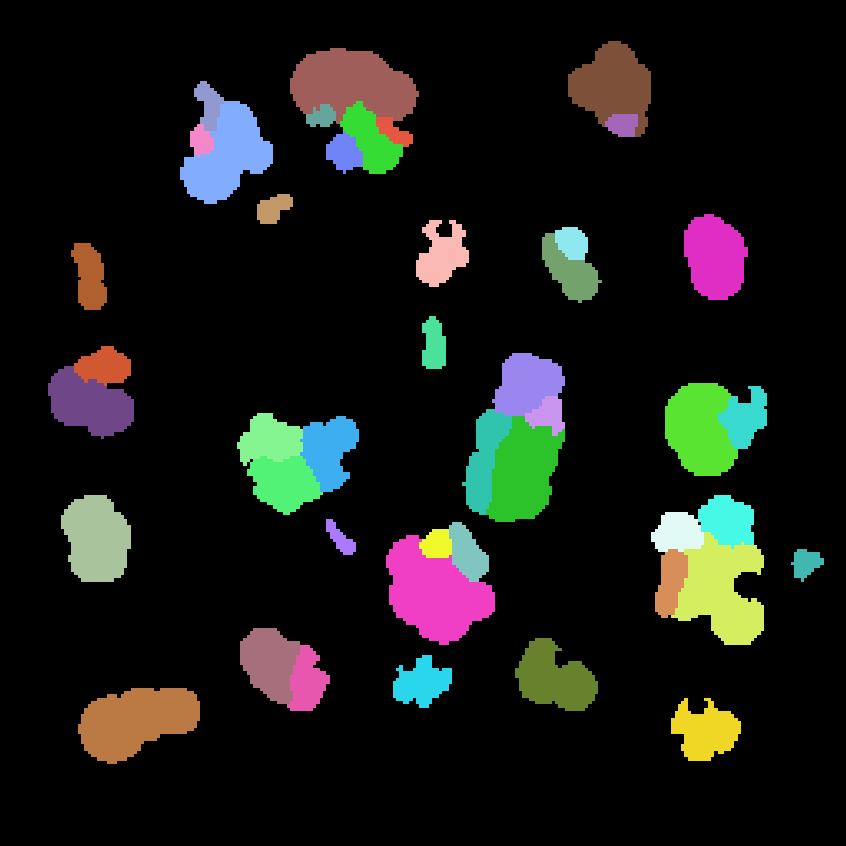} 
\put(2,2){\textcolor{white}{\footnotesize \bfseries (b)}} 
\end{overpic}
} \hfill
\subfigure{
\begin{overpic}[width=0.3\columnwidth]{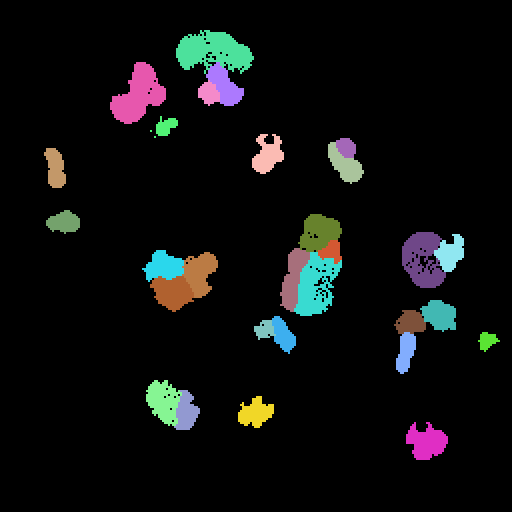} 
\put(2,2){\textcolor{white}{\footnotesize \bfseries (c)}} 
\end{overpic}
} \hfill
\subfigure{
\begin{overpic}[width=0.3\columnwidth]{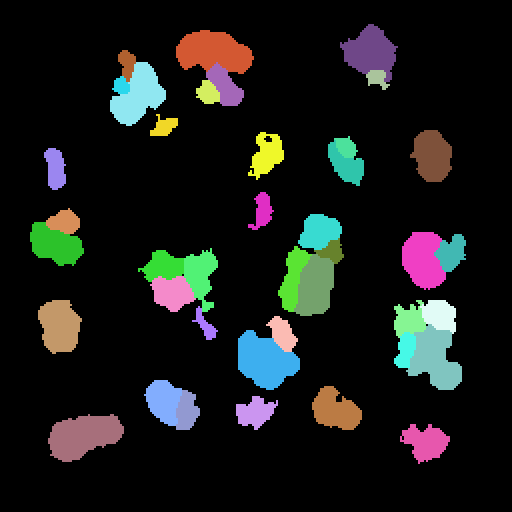} 
\put(2,2){\textcolor{white}{\footnotesize \bfseries (d)}} 
\end{overpic}
} \hfill
\subfigure{
\begin{overpic}[width=0.3\columnwidth]{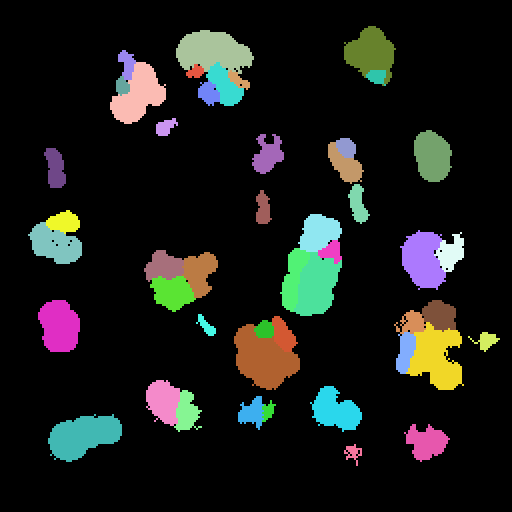} 
\put(2,2){\textcolor{white}{\footnotesize \bfseries (e)}} 
\end{overpic}
} \hfill
\subfigure{
\begin{overpic}[width=0.3\columnwidth]{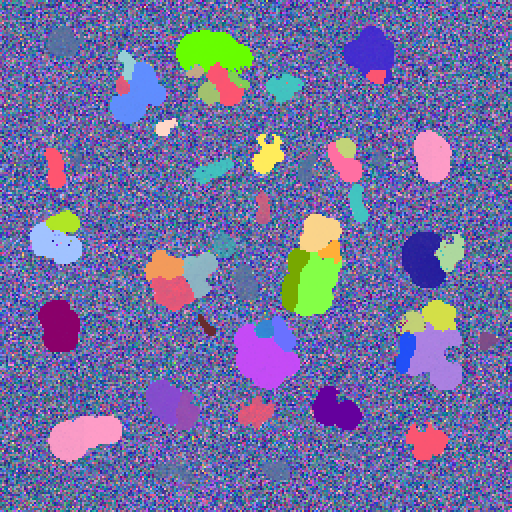} 
\put(2,2){\textcolor{white}{\footnotesize \bfseries (f)}} 
\end{overpic}
} \hspace*{\fill} 
\vspace{-6pt}
\caption{Results for synthetic data. 
(a) Maximum projection of the image sequence; 
(b) ground-truth blinking-cell supports, excluding static distractors; 
extracted supports from (c) Suite2p~\cite{stringer2026suite2p}, 
(d) CaImAn~\cite{giovannucci2019caiman}, and (e) PM-Adaptive; 
and (f) spatial visualization of the PM-Adaptive embedding.
In (b)--(e), different colors denote distinct source supports; 
in (f), the three latent coordinates are mapped to RGB for visualization.}
\label{fig:syn_results}
\vspace{-9pt}
\end{figure*}

\myparagraph{Data.}
We evaluated the method on a synthetic dataset and an in vivo calcium imaging 
sequence. 
The synthetic data consists of a sequence of 44 irregular cells with 180 frames 
of 256$\times$256 pixels, including isolated and touching cells, heterogeneous 
shapes and fluorescence patterns, asynchronous calcium events, static distractors, 
and multiple noise sources (\Fig~\ref{fig:syn_results}(a,b)). 
The in vivo data was acquired using a multiphoton microscope from an awake, 
head-fixed NMRI-nude mouse with a chronic cranial 
window~\cite{osswald2015,hausmann2023} and implanted BG5 glioblastoma cells 
expressing GCaMP7b and tdTomato~\cite{venkataramani2022}.
The image sequence consists of 980 frames of 512$\times$512 pixels with 
a spatial resolution of \qty{0.59}{\micro\metre} per pixel and a frame interval 
of \qty{0.94}{\second} (\Fig~\ref{fig:img_umap_pm}(a,b)). 
We employed the GCaMP7b (green) channel for analysis. 

\myparagraph{Implementation details.}
In all experiments, we set $D\!=\!3$ to enable direct RGB visualization of the latent 
coordinates.
For PM optimization, we employed a $64\!\times\!64\!\times\!64$ mesh with bounds 
fixed from the spectral initialization and a 75\% margin on each side. 
The optimization used $\beta\!=\!4$ and ran for 120 iterations with a linearly 
decaying step size $\eta_t\!=\!0.1(1\!-\!t/120)$.

\begin{table}[t]
\vspace{-3pt}
\centering
\caption{Performance for the synthetic dataset.}
\small
\setlength\tabcolsep{8pt}
\begin{tabular}{l|cccc}
\hline
\rule{0pt}{2.5ex} & \multicolumn{2}{c}{All cells}
& \multicolumn{2}{c}{Touching cells} \\
Method & F1 $\uparrow$ & Recall $\uparrow$ & F1 $\uparrow$ & Recall $\uparrow$ \\
\hline
\rule{0pt}{2.5ex}Suite2p \cite{stringer2026suite2p} & 0.795 & 0.659 & 0.836 & 0.719 \\
CaImAn \cite{giovannucci2019caiman} & 0.952 & 0.909 & 0.951 & 0.906 \\
PM-Euclidean & 0.946 & 1.000 & 0.985 & 1.000 \\
PM-Cosine & 0.966 & 0.955 & 0.952 & 0.938 \\
PM-Adaptive & 0.967 & 1.000 & 1.000 & 1.000 \\
\hline
\end{tabular}
\label{tab:synthetic_results}
\vspace{-6pt}
\end{table}

\myparagraph{Results on synthetic data.}
We studied three variants of our method using Euclidean, cosine, and adaptive fused 
temporal graphs (PM-Euclidean, PM-Cosine, and PM-Adaptive) and performed a 
comparison with Suite2p~\cite{stringer2026suite2p} and 
CaImAn~\cite{giovannucci2019caiman}. 
Predicted and ground-truth source supports were matched one-to-one by the 
Hungarian algorithm using an IoU threshold of 0.30. 
We report F1 score and recall for all cells and for the subset of touching cells. 
Recall quantifies source-recovery completeness, whereas F1 additionally penalizes 
false positives.

As shown in \Tab~\ref{tab:synthetic_results}, PM-Adaptive achieved the best 
overall F1 score while recovering all ground-truth cells. 
PM-Euclidean also reached perfect recall (1.000), but its lower F1 score indicates 
more false positives.
PM-Cosine reduced false positives, but missed several cells. 
PM-Adaptive avoids this trade-off by combining the complementary information 
captured by the Euclidean and cosine temporal graphs. 
The advantage was most pronounced for touching cells, where PM-Adaptive 
achieved 1.000 for both F1 and recall, substantially outperforming CaImAn 
and Suite2p.

The visual results in \Fig~\ref{fig:syn_results} are consistent with these  
quantitative results. 
Suite2p produces a visibly incomplete set of source supports, while CaImAn 
recovers most cells but shows remaining errors in crowded regions. 
PM-Adaptive produces source supports that more closely match the ground 
truth, including those for touching cells. 
In addition, the spatial visualization of the PM-Adaptive embedding shows 
distinct latent representations across cell regions.
The latent representation is subsequently decoded into the extracted supports.

\begin{figure}[t]
\centering
\subfigcapskip=-2pt
\subfigure[CaImAn~\cite{giovannucci2019caiman}]{
\includegraphics[width=0.389\columnwidth]{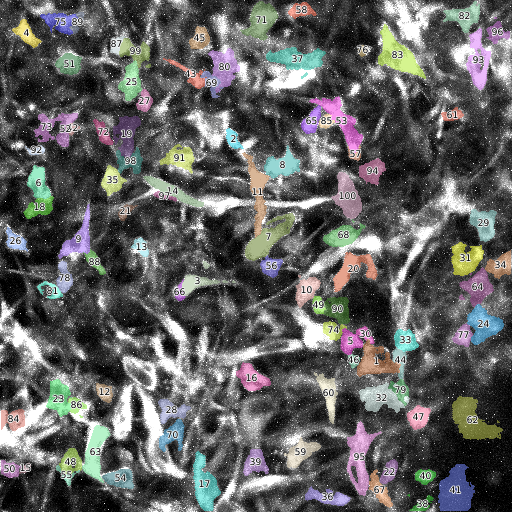}} \hspace{-4pt}
\subfigure[PM-Adaptive]{ 
\label{fig:real_results_b}
\includegraphics[width=0.389\columnwidth]{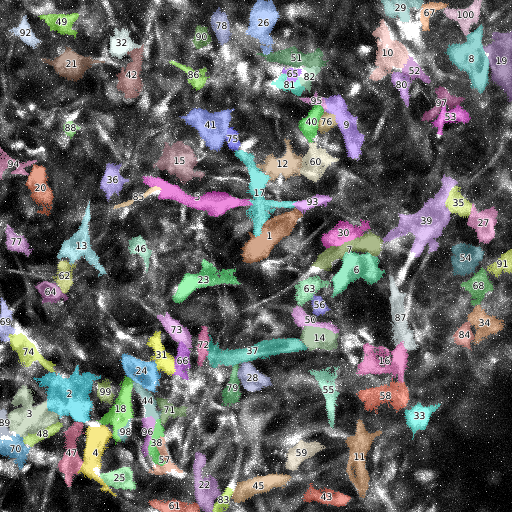}} \hspace{-6pt}
\subfigure[Traces]{ 
\label{fig:real_results_c}
\includegraphics[width=0.179\columnwidth]{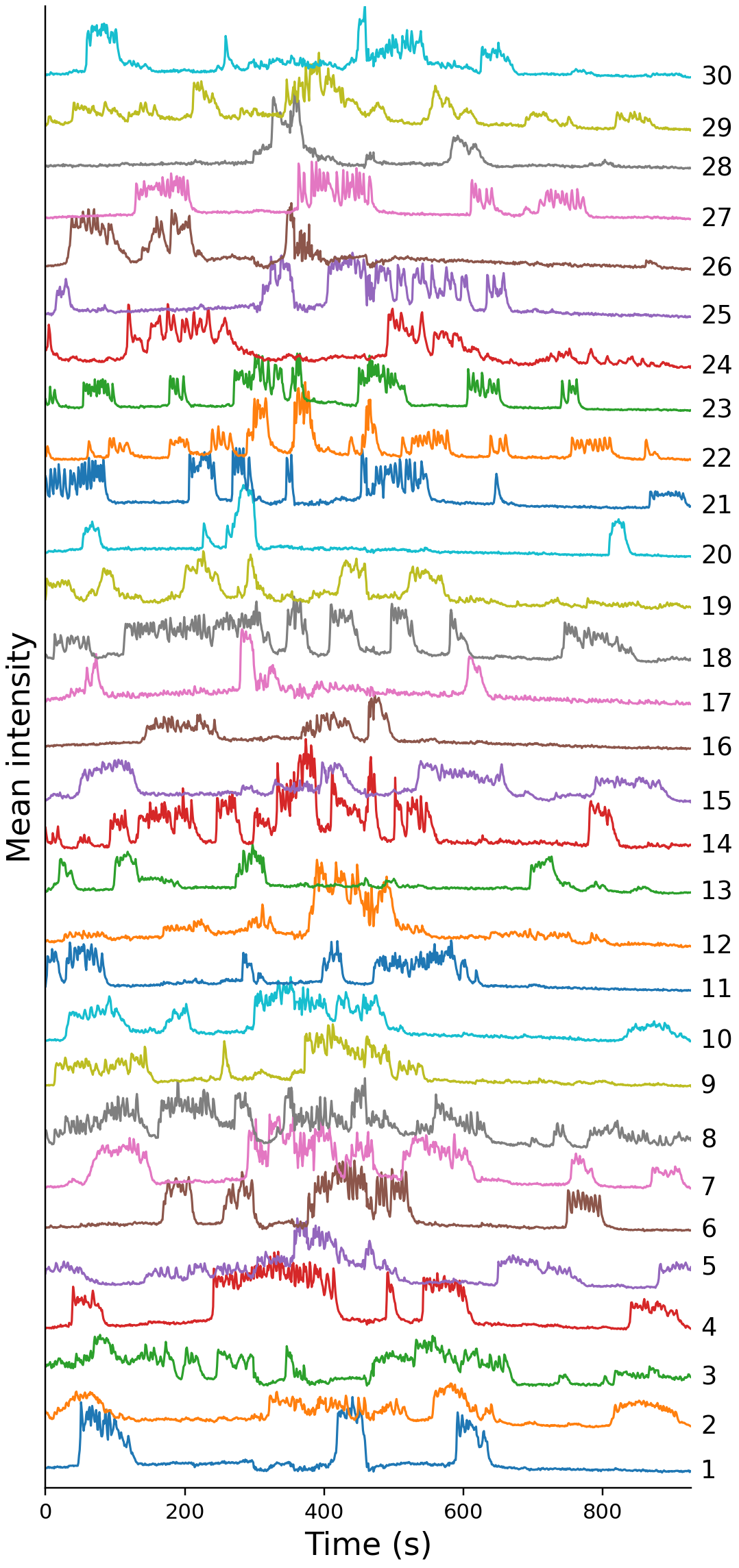}}
\vspace{-9pt}
\caption{Results for real calcium imaging data. 
Filled supports are overlaid on the maximum projection of the image sequence. 
(a) Top 100 accepted CaImAn components ranked by trace SNR. 
(b) Top 100 PM-Adaptive supports ranked by the ACF activity feature. 
(c) Mean fluorescence traces of the top 30 PM-Adaptive supports, with 
IDs corresponding to those in~(b). 
Traces are vertically offset for visualization.}
\label{fig:real_results}
\vspace{-3pt}
\end{figure}

\myparagraph{Results on real calcium imaging data.}
We further evaluated the method qualitatively on the in vivo calcium imaging 
sequence. Since exhaustive manual annotation of all active cells is not available, 
we focus on the spatial supports and the corresponding extracted fluorescence 
traces. 
Compared with Suite2p, CaImAn produced a more interpretable result and is 
therefore shown as the baseline in \Fig~\ref{fig:real_results}. 
For visualization, we display the top 100 accepted CaImAn components ranked 
by trace SNR and the top 100 PM-Adaptive supports ranked by the ACF activity 
feature.

While both methods identify numerous fluorescent structures, PM-Adaptive 
produces spatially coherent supports for a wide range of irregular, elongated, 
and closely adjacent cell morphologies (\Fig~\ref{fig:real_results_b}). 
This behavior is also reflected in the PM-Adaptive embedding (\Fig~\ref{fig:pm}), 
where neighboring cellular structures exhibit distinct latent representations. 
This can be attributed to the fact that the proposed method relies primarily on 
temporal similarity and does not impose an explicit cellular shape prior during 
source discovery. 
The extracted traces of the top-ranked PM-Adaptive supports show distinct 
transient and sustained fluorescence dynamics across cells. 
The support IDs in \Fig~\ref{fig:real_results_b} correspond directly to those in 
the trace plot in \Fig~\ref{fig:real_results_c}, illustrating the spatial origin of the 
recovered temporal signals.

\myparagraph{Computational efficiency.}
Using the same adaptive fuzzy graph and spectral initialization, we compared 
the time for embedding optimization for the real calcium imaging data.
On a MacBook Pro with an Apple M1 Pro and 16 GB unified memory, UMAP 
required \qty{148.3}{\second}, whereas PM-Adaptive required 
\qty{22.8}{\second} and was about 6.5 times faster.  
This gain was obtained with a CPU implementation.
The particle-mesh operations are highly parallelizable and may benefit further 
from a GPU implementation.

\section{Summary}
\vspace{-3pt}

We presented a deterministic temporal embedding framework for calcium imaging 
trace extraction that combines adaptive Euclidean--cosine fuzzy graphs with 
particle-mesh optimization of a class-balanced cross-entropy objective.
The method separates temporally distinct sources in a low-dimensional latent space 
and converts the resulting spatially coherent regions into reliable cellular supports 
without imposing an explicit cellular shape prior. 
Experiments on synthetic and in vivo data demonstrated robust recovery of irregular 
and spatially touching sources together with efficient deterministic optimization. 
The strong downstream performance obtained with a simple spatial decoder further 
suggests that source separation is achieved primarily in the temporal embedding 
rather than through complex spatial post-processing.

\newpage

\myparagraph{Acknowledgments:} 
This work was supported by the 2024 AACR-Novocure Cancer Research Grant, 
Grant Number 24-60-62-RATL awarded to MR.
The authors declare no relevant conflicts of interest.
ChatGPT (OpenAI) was used for language refinement and assistance with coding 
for the experiments.

\vspace{9pt}
\myparagraph{Compliance with ethical standards:} 
The animal experiments underlying the imaging data were approved by the 
Regierungspräsidium Karlsruhe (G110/21).
This work is a study for which no ethical approval was required.

\bibliographystyle{IEEEtran}
\bibliography{bib_medium,bib_loc}

\end{document}